\documentclass[conference]{IEEEtran}
\IEEEoverridecommandlockouts

\usepackage{comment}
\usepackage{caption}
\usepackage{stfloats} 
\usepackage{booktabs}
\usepackage{diagbox}
\usepackage{tabularx}
\usepackage{multirow} 
\usepackage{pifont}

\usepackage{varwidth}
\usepackage{amsmath,amssymb,amsfonts,graphicx}
\usepackage{makecell}
\usepackage{hyperref}
\usepackage{cite}
\usepackage{graphicx,epstopdf,caption,url}  
\usepackage{amssymb} 
\usepackage{array}   
\usepackage{arydshln} 
\usepackage{graphics}
\usepackage{color}   
\usepackage{graphicx}

\begin{document}
%
\title{Multimodal Large Language Models: A Survey}

\author{Jiayang Wu$^{1}$, Wensheng Gan$^{1,2*}$\thanks{\IEEEauthorrefmark{1}Corresponding author: wsgan001@gmail.com. Please cite: J. Wu, W. Gan, Z. Chen, S. Wan, and P. S. Yu, “Multimodal Large Language Models: A Survey,” in \textit{IEEE International Conference on Big Data}, pp. 1–10, 2023.}, Zefeng Chen$^{1}$, Shicheng Wan$^{3}$, Philip S. Yu$^{4}$ \\ \\

 	$ ^{1} $Jinan University, Guangzhou 510632, China\\ 
 	$ ^{2} $Pazhou Lab, Guangzhou 510330, China\\
     $ ^{3} $South China University of Technology, Guangzhou 510641, China\\
     $ ^{4} $The University of Illinois Chicago, Chicago, USA \\
}
\maketitle

\begin{abstract}
  The exploration of multimodal language models integrates multiple data types, such as images, text, language, audio, and other heterogeneity. While the latest large language models excel in text-based tasks, they often struggle to understand and process other data types. Multimodal models address this limitation by combining various modalities, enabling a more comprehensive understanding of diverse data. This paper begins by defining the concept of multimodal and examining the historical development of multimodal algorithms. Furthermore, we introduce a range of multimodal products, focusing on the efforts of major technology companies. A practical guide is provided, offering insights into the technical aspects of multimodal models. Moreover, we present a compilation of the latest algorithms and commonly used datasets, providing researchers with valuable resources for experimentation and evaluation. Lastly, we explore the applications of multimodal models and discuss the challenges associated with their development. By addressing these aspects, this paper aims to facilitate a deeper understanding of multimodal models and their potentiality in various domains.
\end{abstract}

\begin{IEEEkeywords}
   modalities, language models, multimodal models, large models, survey.
\end{IEEEkeywords}

\IEEEpeerreviewmaketitle

\section{Introduction}

A multimodal model combines multiple data types, including images, text, audio, and more. Traditional large language models (LLMs) \cite{gan2023large,gan2023model} are primarily trained and applied to text data, but they have limitations in understanding other data types. Pure text LLMs, such as GPT-3 \cite{dale2021gpt}, BERT \cite{devlin2018bert}, and RoBERTa \cite{liu2019roberta}, excel in tasks like text generation and encoding, but they lack a comprehensive understanding and processing of other data types. To address this issue, multimodal LLMs integrate multiple data types, overcoming the limitations of pure text models and opening up possibilities for handling diverse data types. GPT-4 \cite{sanderson2023gpt} serves as an excellent example of a multimodal LLM. It can accept inputs in the form of both images and text, and it demonstrates human-level performance in various benchmark tests. Multimodal perception is a fundamental component for achieving general artificial intelligence, as it is crucial for knowledge acquisition and interaction with the real world. Furthermore, the application of multimodal inputs greatly expands the potential of language models in high-value domains, such as multimodal robotics, document intelligence, and robot technology. Research indicates that native support for multimodal perception provides new opportunities for applying multimodal LLMs to novel tasks. Through extensive experimentation, multimodal LLMs have shown superior performance in common-sense reasoning compared to single-modality models, highlighting the benefits of cross-modal transfer for knowledge acquisition.

In recent years, the development of multimodal models has showcased additional application possibilities. Apart from text generation models, multimodal models have been increasingly applied in fields such as human-computer interaction, robot control, image search, and speech generation. However, transferring the capabilities of LLMs to the domain of multimodal text and images remains an active area of research, as pure-text LLMs are typically trained only on textual corpora and lack perceptual abilities for visual signals. There are several reviews for multimodal models, but each of these articles has a different focus. Summaira \textit{et al.} \cite{summaira2022review} provided a detailed introduction to the application of different modalities by categorizing them based on modes. Wang \textit{et al.} \cite{wang2023large} presented a comprehensive compilation of the latest algorithms used in multimodal large-scale models and the datasets employed in recent experiments, offering convenience to readers. Yin \textit{et al.} \cite{yin2023survey} classified and differentiated various types of multimodal algorithms in recent years within their review. 

However, these articles primarily start with an introduction to large-scale models, lacking an overview of the development process and practical applications of multimodal models. This paper aims to address this gap by starting with the fundamental definition of multimodal. It provides an overview of the historical development of multimodal algorithms and discusses the potential applications and challenges in this field.

\begin{itemize}
    \item  We start by defining the concept of multimodal models/algorithms, and then delve into the historical development of multimodal algorithms.
    
    \item We provide a practical guide for various technical aspects related to multimodal models, including knowledge representation, learning objective selection, model construction, information fusion, and the usage of prompts.
    
    \item We review the up-to-date algorithms used in multimodal models, along with commonly used datasets. This provides basic resources for future research and evaluation.
    
    \item Finally, we explore several applications of multimodal models and discuss several key challenges that arise from their current development. 
\end{itemize}

The rest of this article is organized as follows: In Section \ref{sec:relatedwork}, we discuss related concepts of the multimodal. In Section \ref{sec:technical}, we indicate the practical guide for technical points. Moreover, in Section \ref{sec:model}, we organized relevant models. Moreover, we present several promising directions for multimodal and various types of datasets in Section \ref{sec:tasks} and highlight the challenges in Section \ref{sec:challenges}. Finally, we conclude this survey in Section \ref{sec:conclusion}.

\section{Related Concepts} \label{sec:relatedwork}

\begin{figure}[ht]
    \centering
    \includegraphics[clip,scale=0.25]{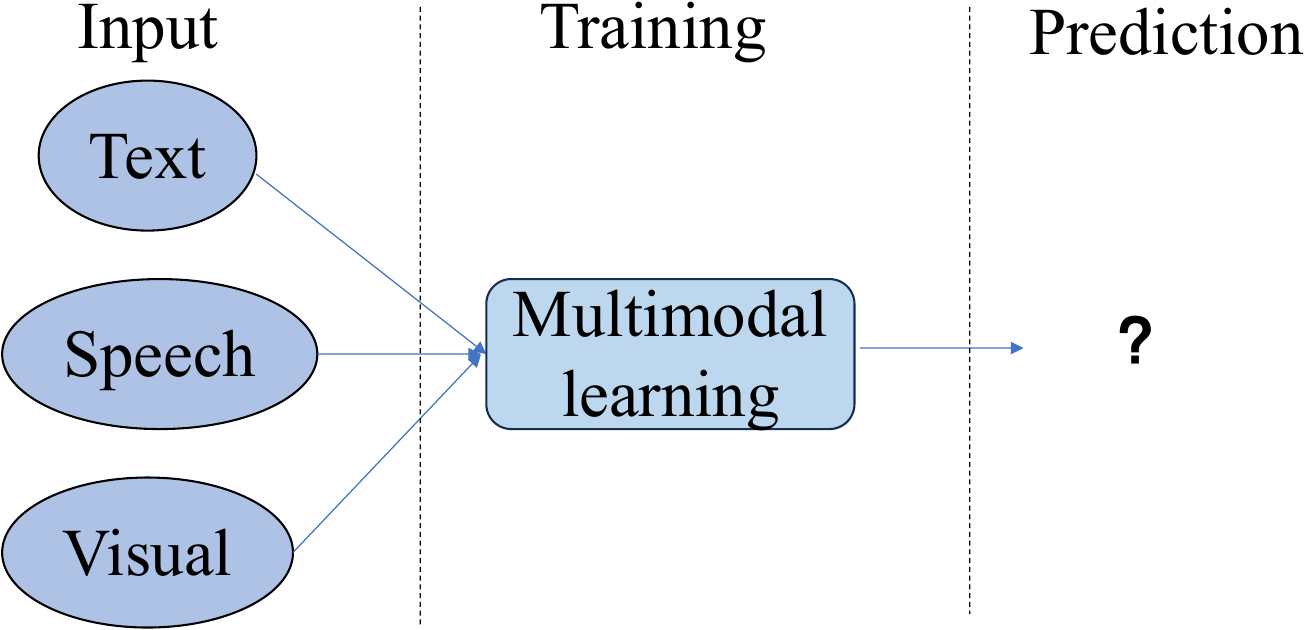}
    \caption{The definition of multimodal.}
    \label{fig:types_of_data}
\end{figure}

Multimodal refers to expressing or perceiving complex things through multiple modalities, as shown in Fig. \ref{fig:types_of_data}. 

Multi-modality can be classified into homogeneous modalities, such as images captured from two different cameras, and heterogeneous modalities, such as the relationship between images and textual language. Multimodal data, from a semantic perception standpoint, refers to the integration of information from various sensory modalities, such as visual, auditory, tactile, and olfactory inputs, to form a unified and meaningful representation of the environment \cite{turk2014multimodal}. From a data perspective, multimodal data can be seen as a combination of different data types, such as images, numerical data, text, symbols, audio, time series, or complex data structures composed of sets, trees, graphs, and even combinations of various information resources from different databases or knowledge bases. The exploration and analysis of heterogeneous data sources can be understood as multimodal learning. Using multimodal data allows for a more comprehensive and holistic representation of things, making multimodal research an important area of study. Significant breakthroughs have been achieved in areas such as sentiment analysis, machine translation, natural language processing, and cutting-edge biomedical research \cite{ortega2009multiscenario} by leveraging multimodal approaches.

\begin{figure}[ht]
    \centering
    \includegraphics[clip,scale=0.27]{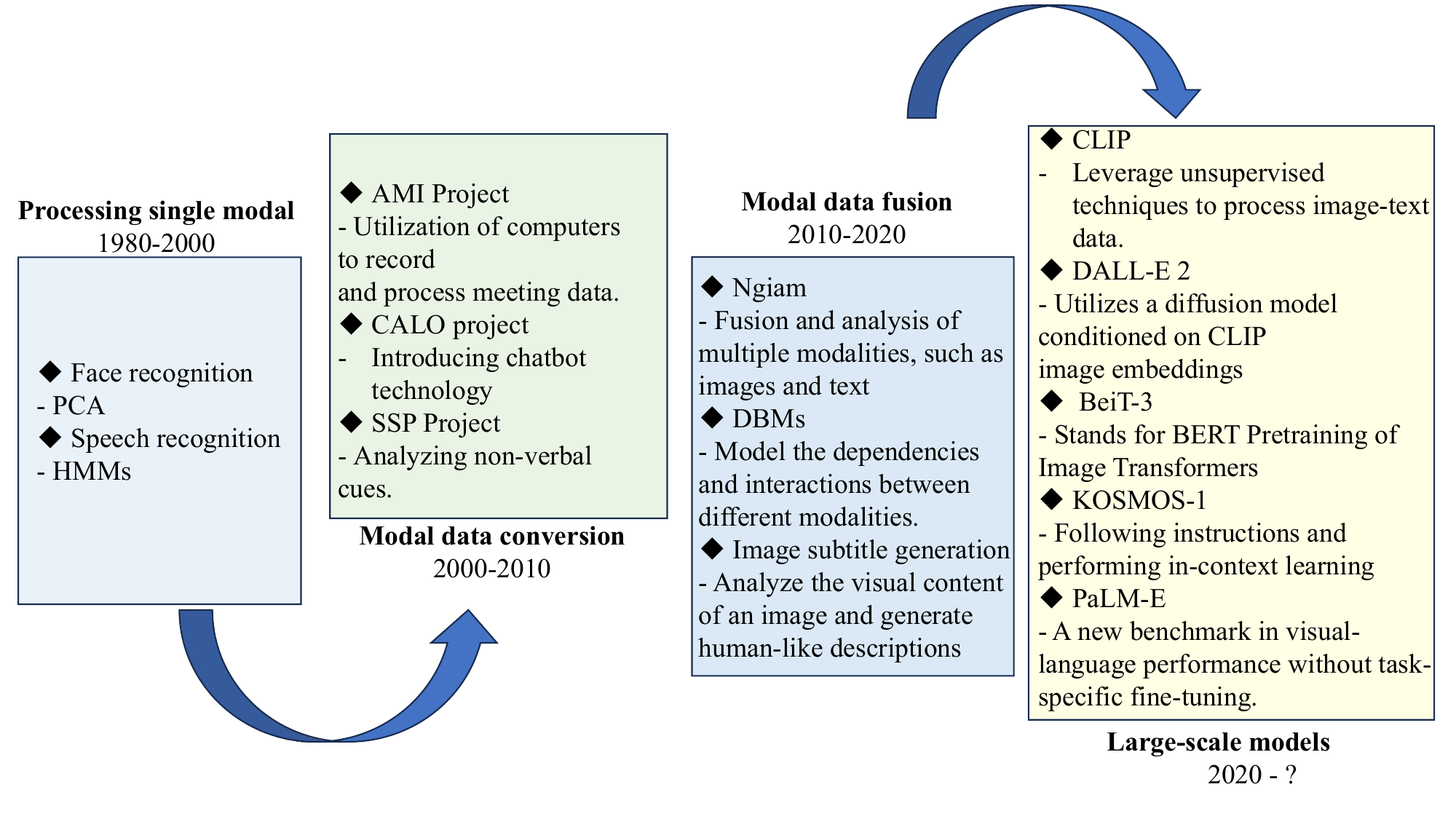}
    \caption{Four distinct stages of multimodal research.}
    \label{fig:four_stage}
\end{figure}

During the evolution of multimodal research, four distinct stages can be identified, as shown in Fig.\ref{fig:four_stage}. 

\textbf{Single modality (1980-2000)}. It was characterized by its reliance on basic computing capabilities. In the 1980s, statistical algorithms and image-processing techniques were used for face recognition systems. The work laid the foundation for early methods in this field. Concurrently, the research team at IBM made significant contributions to speech recognition, e.g., the use of hidden Markov models (HMMs) \cite{bahl1986maximum}, which improved the accuracy and reliability of speech recognition technology. Further progress was made in the 1990s. Kanade's team developed the Eigenfaces method for face recognition \cite{satoh1997name}. This utilized principal component analysis (PCA) to extract facial features and recognize individuals based on statistical patterns in face images \cite{lien1998automated}. Companies such as Dragon Systems focused on advancing speech recognition systems, developing technology capable of converting spoken language into written text with increasing accuracy \cite{larocca1999path}.

\textbf{Modality conversion (2000-2010)}. In this stage, researchers devoted significant resources to the study of human-computer interaction. The goal was to enable computers to simulate human behavior and enhance convenience in the daily lives of people. Several notable advancements took place during this period. In 2001, the AMI project proposed the utilization of computers to record and process meeting data. This project aimed to develop technologies that could analyze audio, video, and text data from meetings, enabling more efficient information retrieval and collaboration \cite{carletta2005ami}. In 2003, the CALO project made significant contributions by introducing chatbot technology, which served as the predecessor to Siri. The CALO project, which stands for ``Cognitive Assistant that Learns and Organizes", aimed to develop an intelligent virtual assistant capable of understanding and responding to human language and performing tasks \cite{tur2010calo}. In 2008, the social signal processing (SSP) project introduced the concept of social signal processing networks. This project focused on analyzing non-verbal cues, such as facial expressions, gestures, and voice tones, to understand social interactions and facilitate more natural human-computer communication \cite{vinciarelli2008social}.

\textbf{Modality fusion (2010-2020)}. In this stage, the integration of deep learning techniques and neural networks led to notable advancements in the field. In 2011, a pioneering multimodal deep learning algorithm was introduced by Ngiam \cite{ngiam2011multimodal}. This algorithm played a crucial role in advancing the field by enabling the fusion and analysis of multiple modalities, such as images and text. It facilitated the joint learning of features from different modalities and contributed to enhanced performance in tasks like image classification, speech recognition, and video analysis. In 2012, a multimodal learning algorithm based on deep Boltzmann machines (DBMs) \cite{hinton2012better} aimed to model the dependencies and interactions between different modalities. By leveraging the power of deep learning and the generative modeling capabilities of DBMs, we can capture the intricate relationships among modalities and improve the understanding and representation of complex multimodal data. In 2016, a neural image captioning algorithm with semantic attention was introduced \cite{you2016image}, revolutionizing the way images were processed and described. This algorithm had the functionality to generate descriptive captions for images, allowing automated image understanding and interpretation. By combining computer vision techniques with deep neural networks, the algorithm could analyze the visual content of an image and generate human-like descriptions, improving accessibility and enabling applications like automatic image tagging, image search, and assistive technologies for the visually impaired.

\textbf{Large-scale multimodal (2020-?)}. The rapid development of large-scale models has opened up new opportunities for multimodal algorithms. In 2021, the CLIP model was introduced \cite{radford2021learning}. By shattering the conventional paradigm of fixed category labels, CLIP liberates the burden of assembling massive datasets with predetermined class counts. Instead, CLIP empowers the collection of image-text pairs and leverages unsupervised techniques to either predict their similarity or generate them. In 2022, DALL-E 2, a product in OpenAI, utilizes a diffusion model conditioned on CLIP image embeddings \cite{ramesh2022hierarchical}. It can generate high-quality images and artwork based on text prompts. Microsoft also introduced BEiT-3 (BERT Pretraining of Image Transformers) \cite{bao2021BEiT}. BEiT-3 uses a shared multiway transformer structure to complete pre-training through masked data. It can be migrated to various downstream tasks of vision and visual language. In 2023, KOSMOS-1 was released by Microsoft \cite{huang2023language}. KOSMOS-1 is a cutting-edge multimodal LLM that boasts an impressive array of capabilities, including the ability to process and integrate information from diverse modalities, follow instructions with precision, and adapt to new contexts through in-context learning. This model integrates language and perception to enable itself to see and speak, making it proficient in tasks such as visual dialogue, image captioning, and zero-shot image classification. Another notable model, namely PaLM-E \cite{driess2023palm}, combines advanced language and vision models, e.g., PaLM and ViT-22B. They could excel in visual tasks like object detection and scene classification, while also demonstrating proficiency in language tasks, e.g., generating code and solving math equations. PaLM-E provides a new benchmark in visual-language performance without task-specific fine-tuning.

\section{Practical Guide for Technical Points}\label{sec:technical}

\begin{figure}[ht]
    \centering
    \includegraphics[clip,scale=0.55]{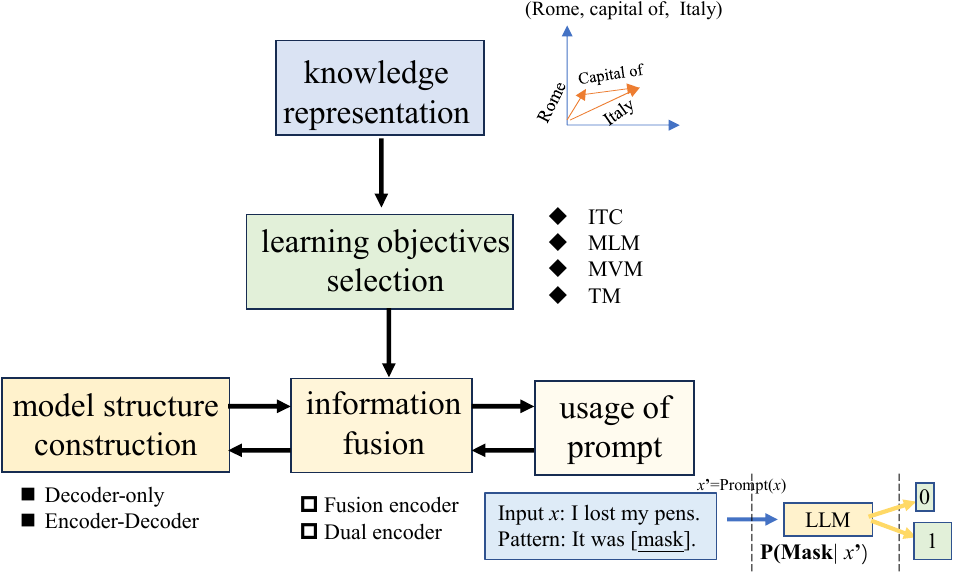}
    \caption{The technical points of multimodal models.}
    \label{fig:techniques}
\end{figure}

The technical points of multimodal large models include, but are not limited to, knowledge representation, learning objective selection, model structure construction, information fusion, and the usage of prompts, as shown in Fig. \ref{fig:techniques}.

\textbf{Knowledge representation.} Both text and images require tokenization and embedding. Tokens are the basic units of input for the models, while embeddings are the vector representations of tokens used for calculations. In the case of text, Word2Vec \cite{mikolov2013efficient} was commonly used for tokenization, including some methods like CBOW and Skip-gram. Although Word2Vec is computationally efficient, it suffers from vocabulary limitations. As a result, subword tokenization methods, such as byte pair encoding \cite{bostrom2020byte}, divide words into smaller units. This approach has been applied to various transformer models, like BERT. In contrast, image tokenization is more complex than text. It can be categorized into three types \cite{yang2022visual}, including region-based, grid-based, and patch-based. Region-based methods utilize pre-trained object detectors to extract features. Grid-based methods directly apply convolutional neural networks to extract grid-based information from images. While patch-based methods involve dividing the image into smaller blocks and extracting linear projections from those blocks. According to the data from the METER model \cite{dou2022empirical}, optimizing the visual feature side has a much greater impact on the results than optimizing the text side. In the construction of multimodal pretraining models, the embedding layers or complexity of visual features surpass those of text features, highlighting the significance of visual information. Multimodal models can learn more knowledge from visual features. 

\textbf{Learning objectives selection.} It is crucial in multimodal pretraining. Currently, common learning tasks in multimodal pretraining include image-text contrast (ITC), masked language modeling (MLM), masked visual modeling (MVM), and image-text matching (TM) \cite{rao2023retrieval}. ITC involves constructing positive and negative sample pairs to align images and texts through contrastive learning. In addition, by leveraging MLM and MVM techniques, it can learn to infer the subtle connections between language and visual data by reconstructing masked linguistic tokens from a combination of linguistic knowledge and visual cues. In this way, it could improve its ability to comprehend and generate multimodal content. TM can be seen as a binary classification task that aims to predict whether an image and text pair match. In general, using different learning objectives in combination can enhance the performance of multimodal models. For instance, in the UNITER model, incorporating more learning objectives generally leads to better results. UNITER utilizes multiple learning objectives, such as MLM and  ITC, and performs well across various specialized scenarios. However, using too many learning objectives may not always yield favorable results. This was validated in the experiment on the METER \cite{dou2022empirical}.

\textbf{Model construction.} Based on the different model structures, multimodal models can be categorized into encoder-only and encoder-decoder models. Encoder-only models utilize only the encoder part of the Transformer. The multimodal input is directly processed by the encoder to produce the output. Common examples of encoder-only models include CLIP \cite{ramesh2022hierarchical} and ALBEF \cite{li2021align}, which are suitable for tasks like image-text retrieval but not ideal for tasks like image captioning. The encoder-decoder models incorporate both the encoder and decoder parts of the Transformer. The decoder receives the previously generated tokens and its own output to generate the output sequence auto-regressively. Encoder-decoder models, such as T5 \cite{raffel2020exploring} and SimVLM \cite{wang2021simvlm}, leverage the decoder's capabilities and are suitable for generation tasks, but may not be as well-suited for tasks like image-text retrieval.

\textbf{Information fusion.} After encoding different modalities separately, it is necessary to design an encoder for multimodal encoding. Based on different fusion methods, multimodal models can be categorized into fusion encoder and dual encoder models \cite{wang2021distilled}. The fusion encoder utilizes fusion methods to interact between modalities. Through self-attention or cross-attention operations, the fusion encoder generates fused representations of the modalities. Fusion methods mainly include single-stream and dual-stream approaches. The single-stream approach assumes that there exists a simple alignment or correlation between the two modalities, and applies self-attention mechanisms directly to the modalities before concatenating them. The dual-stream model assumes that intra-modal and cross-modal interactions should be modeled separately to obtain better multimodal representations using cross-attention mechanisms. Fusion encoders model cross-modal interactions at different levels and have achieved good performance in certain inference tasks. However, in tasks such as image-text retrieval, encoding the interactions of all image-text pairs leads to slow inference speed. The dual encoder employs separate single-modal encoders to encode the two modalities. After sufficient encoding, a simple dot product or shallow attention layer is used to calculate similarity scores between them, without relying on complex Transformer structures. The fusion encoder is suitable for inference tasks, while the Dual encoder is suitable for retrieval tasks. Therefore, we combine different model architectures or information fusion methods to enhance the capabilities of multimodal models. This is also the mechanism behind the implementation of multimodal unification. For example, VLMO adopts the ``Three Experts'' approach, pretraining on image-only, text-only, and image-text data to handle different modalities, and achieves good performance in tasks such as inference and retrieval \cite{bao2022vlmo}.

\textbf{The usage of prompt.} The prompt method is primarily used to reduce the gap between pretraining and fine-tuning in downstream tasks. By modifying the templates of downstream tasks, prompt aims to minimize the differences between pretraining and fine-tuning, thereby reducing the cost of fine-tuning and improving the model's performance in downstream applications. It has the ability to handle zero or small data samples, which has been widely adopted in various LLMs \cite{white2023prompt}. The prompt method plays a crucial role in multimodal pretraining tasks as well. For example, in visual ChatGPT \cite{wu2023visual}, a prompt manager is used to generate informative prompts that facilitate ChatGPT's understanding and generation of related images. In CLIP, the prompt method is applied in zero-shot tasks by generating informative prompts for text, resulting in improved performance \cite{ramesh2022hierarchical}.

\section{Practical Guide for Algorithms}\label{sec:model}

The algorithms in multimodal can be categorized into two types, including foundation models and large-scale multimodal pre-trained models. The foundation modal is the basic framework for multimodal. Many novel large-scale multimodal pre-trained models are improved based on it.

\subsection{Foundation model.}

\textbf{Transformer} \cite{vaswani2017attention} was proposed in 2017, disrupting traditional deep learning models and achieving good performance in machine translation tasks. It gained attention for its ability to undergo self-supervised pre-training on large-scale corpora and subsequent fine-tuning on downstream tasks. This paradigm has been followed by many pre-trained large-scale models. The weight-sharing property of the Transformer, which is independent of the input sequence length, makes it suitable for multimodal applications. Certain modules within the model can share weight parameters. The weight-sharing concept in the Transformer arises from the fact that both the self-attention module and the feed-forward neural network are unaffected by the length of the input sequence. This weight-sharing concept can also be applied to multimodal models. For example, in a multimodal setting involving images and text, the weight parameters learned from image training can be used for text training, and the results remain effective, sometimes even without the need for fine-tuning.

\textbf{VIT.} The exceptional performance of the Transformer model with its self-attention mechanism in the domain of natural language processing (NLP) has attracted much attention in computer vision. Many studies have started to incorporate the Transformer mechanism into computer vision tasks. However, the Transformer has limitations in terms of input data size, requiring careful consideration of input strategies. Google, drawing inspiration from previous work, proposed the vision transformer (ViT) model, empowered by powerful computational resources. The ViT model addresses the input size limitation by segmenting images into patches (e.g., dividing an image into 16 patches) \cite{dosovitskiy2020image}. These patches are then processed and transformed into inputs that the Transformer can handle through linear mapping. This breakthrough has bridged the gap between computer vision and NLP. ViT not only enables the Transformer to process images but also introduces more efficient image feature extraction strategies compared to previous approaches.

\textbf{BEiT.} If ViT can be seen as the adaptation of the Transformer model in computer vision, then BEiT can be considered as the adaptation of BERT in computer vision \cite{bao2021BEiT}. Generative pre-training is an important method and training objective in self-supervised learning, where the model learns how to generate data without relying on labels or manual annotations. Generative pre-training has achieved significant success in natural language processing. BEiT addresses two key challenges in generative pre-training for computer vision. The first challenge is how to convert image information into discrete tokens similar to NLP. BEiT uses the discrete visual embedding aggregation method to discretize images. The second challenge is how to incorporate image information into the pre-training process effectively. BEiT leverages the well-established ViT structure to process image information. By addressing these two points, BEiT successfully applies the masked language modeling (MLM) and masked image modeling (MIM) methods to the field of computer vision, bringing generative pre-training into the domain of computer vision and enabling large-scale self-supervised pre-training.

\subsection{Large-scale multimodal pre-trained models}

\textbf{Visual ChatGPT} \cite{wu2023visual} incorporates different visual foundation models (VFMs) to handle various visual tasks, such as image understanding and generation. This allows users to send and receive not only languages but also images, enabling complex visual questions and instructions that require the collaboration of multiple AI models with multi-steps. This system also introduces Prompt Manager, which helps leverage VFMs and receive their feedback in an iterative manner. This iterative process continues until the system meets the requirements of users or reaches the ending condition. By injecting visual model information into ChatGPT through prompts, the system aligns visual features with the text space, enhancing the visual understanding and generation capabilities of ChatGPT. Visual ChatGPT has the ability to handle modalities beyond languages and images. While the system initially focuses on languages and images, it opens up possibilities for incorporating other modalities like videos or voices. This flexibility eliminates the need to train a completely new multi-modality model every time a new modality or function is introduced.

\textbf{MM-REACT} \cite{yang2023mm} combines ChatGPT with various visual models to enable multi-modal tasks, primarily demonstrated through the VQA format. In answering questions, ChatGPT utilizes visual models as tools and decides whether to use them based on the specific question. This system shares similarities with previous works that used caption models and language-image models for VQA. In those approaches, the caption model converted images into text, which was then used as evidence by a larger model to generate answers. However, MM-REACT differs in its ability to autonomously decide whether to invoke visual models.

\textbf{Frozen} \cite{tsimpoukelli2021multimodal} introduced the novel concept of employing LLMs in multi-modal in-context learning. The specific approach involves transforming images into embeddings using a visual encoder. These embeddings are then concatenated with the text, creating a combined data format that integrates both modalities. Subsequently, the model uses an autoregressive approach to predict the next token. Throughout the training process, the LLM remains frozen, while the visual encoder is trainable. This allows the final model to retain its language modeling capabilities while acquiring the ability to perform contextual learning in a multi-modal setting.

\textbf{BLIP-2} \cite{li2023blip} adopts a similar approach to Flamingo in encoding images, utilizing a Qformer model to extract image features. The Qformer plays a role analogous to Flamingo's perceiver resampler. This model then facilitates image-text interaction through cross-attention. During training, BLIP-2 freezes both the visual encoder and LLMs and only fine-tunes the Qformer. However, when fine-tuning on specific downstream task datasets, BLIP-2 unlocks the visual encoder and fine-tunes it alongside Qformer. The training process for BLIP-2 consists of two stages. i) Only Qformer and the visual encoder participate in training. They are trained using classic multi-modal pretraining tasks such as image-text matching, contrastive learning, and image-grounded text generation. This stage enables Qformer to learn how to quickly extract text-related features from the visual encoder. ii) The Qformer-encoded vectors are inserted into the LLM for caption generation. BLIP-2 demonstrates promising performance in both zero-shot and fine-tuning scenarios for VQA. It has good transferability across different datasets for the same task.

\textbf{LLaMA-Adapter} \cite{zhang2023llama} introduces efficient fine-tuning in LLaMA by inserting adapters, which can be extended to multi-modal scenarios. Adapters are adaptation prompt vectors that are concatenated to the last layers of the Transformer as tunable parameters. When applied to multi-modal settings, images are first encoded into multiscale feature vectors using a frozen visual encoder. These vectors are then aggregated through concatenation and projection operations before being element-wise added to the adaptation prompt vectors.

\textbf{MiniGPT-4} \cite{zhu2023minigpt} is a reproduction of certain functionalities of GPT-4 based on the combination of BLIP-2 and Vicuna. It directly transfers the Qformer and visual encoder from BLIP-2 and freezes them along with LLM, leaving only a linear layer on the visual side for fine-tuning. This compression of tunable parameters results in a model size of 15 M. Additionally, a two-stage fine-tuning strategy is adopted. i) Caption generation is used as the training task. The model generates multiple captions, and then these captions are rewritten using ChatGPT to create detailed and vivid descriptions. ii) A set of high-quality image-text pairs is constructed for further fine-tuning. This set of image-text pairs is used to refine the model.

\textbf{LLaVA} \cite{liu2023visual} and MiniGPT-4 are similar, as both aim to achieve multimodal instruction fine-tuning. However, they differ in terms of data generation and training strategies, leading to the development of the LLaVA model. In data generation, LLaVA leverages GPT-4  to create diverse instruction fine-tuning data, including multi-turn QA, image descriptions, and complex reasoning tasks. This ensures that the model is capable of handling a wide range of queries. Since the current interface of GPT-4 only accepts text inputs, image information needs to be transformed into textual format. This study uses the five captions and bounding box coordinates provided for each image in the COCO dataset as textual descriptions inputted to GPT-4. Regarding the training strategy, LLaVA adopts a two-stage approach. i) The model is fine-tuned using 600,000 image-text pairs filtered from the cc3m dataset according to specific rules. The fine-tuning process freezes the visual and language models, focusing only on fine-tuning the linear layer. ii) Using the aforementioned data generation strategy, 160,000 instruction fine-tuning data samples are generated. The model is then further fine-tuned using language model loss. During this stage, the visual model is frozen, and both the linear layer and the language model are fine-tuned.

\textbf{PICa} \cite{yang2022empirical} was the first attempt to use LLMs for solving the VQA task. Its objective was to enable LLM to understand and process image information. To achieve this, previous research employed a caption model to convert images into corresponding textual descriptions. The caption, along with the question, was then inputted into GPT-3, forming a triplet (question, caption, answer), and in-context learning was utilized to train GPT-3 to answer new questions. In the few-shot in-context learning scenario, PICa achieved better performance than Frozen but still fell short of Flamingo. This can be attributed to the loss of visual information during the conversion of images into captions. Visual information plays a crucial role in answering questions, and the process of converting images into text inevitably leads to a loss of visual details and semantics, limiting the model's performance.

\textbf{PNP-VQA} \cite{tiong2022plug} utilizes a caption model and pre-trained language model (PLM) to address the VQA task. However, it differs from PICa in terms of the choice of PLM, as it employs a question-answering model called UnifiedQAv2. PNP-VQA focuses on achieving zero-shot VQA capability. To address the issue of losing image information in captions, PNP-VQA introduces an image-question matching module before generating the captions. This module identifies patches in the image that are most relevant to the given question. Captions are then generated specifically for these selected patches. These caption-patch pairs, along with the original question, are used as context and fed into the UnifiedQAv2 model. This approach ensures that the generated captions are closely related to the question by incorporating relevant image patches as context. By incorporating the Image-Question Matching module and utilizing UnifiedQAv2 as the PLM, PNP-VQA aims to improve the relevance and accuracy of the generated captions for VQA. This strategy allows the model to effectively leverage both image and question information in order to generate more contextually relevant answers.

\textbf{Img2LLM}  \cite{guo2022images} aims to address two main challenges when using LLM for VQA tasks. i) Modality disconnection, where LLM cannot handle visual information effectively; ii) Task disconnection, where LLM, pre-trained through text generation, struggles to utilize captions for VQA without fine-tuning. To overcome these challenges, the authors propose transferring visual information through (question, answer) pairs. Specifically, the approach involves generating captions for images using a caption model or a method similar to PNP-VQA. From these captions, relevant words, such as nouns and adjectives, that could potentially serve as answers to certain questions are extracted. Subsequently, a question generation model is used to generate corresponding questions, thus creating (question, answer) pairs. These pairs serve as demonstrations in in-context learning, aiding LLM in answering questions about the given image. By transmitting visual information through (question, answer) pairs, Img2LLM addresses modality disconnection and task disconnection issues, enabling LLM to better utilize visual information for VQA tasks.

\begin{table*}[ht]
    \caption{The multimodal models}
    \label{table2}
    \renewcommand{\arraystretch}{1.35}
	\begin{tabularx}{\textwidth}{m{1.7cm}<{\centering}m{1.5cm}<{\centering}m{5.3cm}<{\raggedright}m{4.4cm}<{\raggedright}m{0.8cm}<{\centering}m{2.2cm}<{\centering}}
    \toprule
    \hline
    \textbf{Model}  & \textbf{Year}  & \multicolumn{1}{c}{\textbf{Technical points}} & \multicolumn{1}{c}{\textbf{Function}} & \textbf{Paper} & \textbf{Open-source}  \\
    \hline
    Transformer & 2017 & Self-attention, positional Encoding and multi-Head Attention  & Machine translation& \cite{vaswani2017attention} & \checkmark  \\
    \hline
     ViT & 2020  & Patch-based Representation, linear Projection, transformer Encoder & Image classification and generation & \cite{dosovitskiy2020image} & \checkmark \\
    \hline
     BEiT & 2021 & Discrete visual embedding aggregation and MLM & Image Understanding and transfer learning & \cite{bao2021BEiT} & \checkmark \\
    \hline
     VisualChatGPT & 2023 & Invokes multiple VFMs, pre-trained LLMs and prompt management  & Visual queries and instructions & \cite{wu2023visual} & \checkmark \\
    \hline
       MM-REACT & 2023 &  Integration of ChatGPT and vision experts and textual prompt design &  Visual understanding tasks & \cite{yang2023mm} & \checkmark \\
    \hline
      Frozen & 2021 & Few-Shot learning, utilize external knowledge and  soft-prompting philosophy & VQA & \cite{tsimpoukelli2021multimodal} & \checkmark \\
    \hline
     BLIP-2 & 2023 & Pre-trained image encoders, querying Transformer and frozen LLMs & Zero-shot image-to-text generation & \cite{li2023blip} & \checkmark \\
    \hline
     LLaMA-Adapter & 2023 & Fine-tuning instruction-following, learnable adaption prompts and zero-initialized attention mechanism & Vision and language tasks & \cite{zhang2023llama} &  \checkmark \\
    \hline
     MiniGPT-4 & 2023 &  Frozen visual encoder with a frozen LLM, one projection layer and trained by image-text pairs   & Identify humorous elements within images and create websites from handwritten drafts & \cite{zhu2023minigpt} &  \checkmark \\
    \hline
    LLaVA & 2023 & Instruction tuning LLMs and end-to-end trained LLMs & Visual and language understanding and multimodal chat abilities  & \cite{liu2023visual} & \checkmark \\
    \hline
    PICa & 2022 & Utilize GPT-3 as an implicit knowledge base and prompt GPT-3 via image captions  & VQA & \cite{yang2022empirical} & \ding{55}  \\
    \hline
    PNP-VQA & 2022 & Image-question matching module, image captioning module, and question answering module  & Vision-language tasks & \cite{tiong2022plug} & \checkmark \\
    \hline
      Img2LLM & 2022 & Zero-shot generalization and without requiring end-to-end training  & VQA  & \cite{guo2022images} & \checkmark  \\
    \hline
    \bottomrule
    \end{tabularx}
\end{table*}

\section{Practical Guide for Various Tasks} \label{sec:tasks}
\textbf{Image captioning.}  Image captioning is a task that involves generating short textual descriptions for given images. It is a multimodal task that deals with multimodal datasets consisting of images and short textual descriptions. Multimodal translation tasks are open-ended and subjective, so the generated content is not unique. The goal of this task is to convert visual representations into textual representations to address the translation challenge. Models that convert visual modalities into text need to capture the semantic information of the images and need to detect key objects, actions, and features of the objects. Moreover, it should infer the relationships between objects in the image. Image captioning can be used to provide textual alternatives for images, which is particularly helpful for blind and visually impaired users \cite{gurari2020captioning}. By generating short textual descriptions, these users can better understand and perceive the content of the images. It provides them with an opportunity to interact with the visual world, enhancing their experience and engagement.

\textbf{Text-to-Image generation.}  Text-to-image generation is indeed one of the most popular applications of multimodal learning. It addresses the challenge of translating text into images. Models such as OpenAI's  DALL-E 2 \cite{ramesh2022hierarchical} and Google's Imagen \cite{saharia2022photorealistic} have made significant breakthroughs in this area, attracting widespread attention. The work of these models can be the inverse process of image captioning. By providing short textual descriptions as prompts, text-to-image models can generate novel images that accurately reflect the semantics of the text. Recently, there has also been an emergence of text-to-video models. These models have a wide range of applications. They can assist in photo editing and graphic design, while also providing inspiration for digital art. They offer users a tool to directly convert text into visual content, driving the development and innovation of the creative industry. The advancements in these technologies offer new possibilities for creating and understanding images.

\textbf{Sign language recognition.} The goal of this task is to recognize sign language gestures and convert them into text. Gestures are captured through cameras. To accurately recognize the gestures, the corresponding audio and both modalities must be aligned. Sign language recognition is a task based on alignment methods, as it requires the model to align the temporal information of the visual, such as video frames, and audio modalities, such as audio waveforms \cite{albanie2020bsl}. This involves aligning the time between video frames and audio waveforms to identify the gestures and their corresponding spoken language. One commonly used open-source dataset for sign language recognition is the RWTH PHOENIX Weather 2014T dataset \cite{forster2014extensions}, which contains video recordings of German sign language from different signers. The dataset provides both visual and audio modalities, making it well-suited for multimodal learning tasks that rely on alignment methods. By aligning the temporal information of the video and audio, models can leverage both visual and audio features for sign language recognition, thereby improving the accuracy and effectiveness of recognition.

\textbf{Emotion recognition.}  While emotion recognition can be performed using only a single-modal dataset, performance can be improved by utilizing multimodal datasets as input. Multimodal inputs can take the form of video, text, and audio or can incorporate sensor data such as brainwave data \cite{zhao2021emotion}. A real-world example is emotion recognition in music. In this task, the model needs to identify the emotional content of music using audio features and lyrics. In such cases, employing a late fusion approach is appropriate, as it combines the predictions of models trained on individual modalities such as audio features and lyrics to generate the final prediction. The DEAM dataset is specifically designed to support research on music emotion recognition and analysis. It includes audio features and lyrics for over 2,000 songs \cite{aljanaki2017developing}. The audio features encompass various descriptors like MFCC, spectral contrast, and rhythm features, while lyrics are represented using techniques like bag-of-words and word embeddings.

\textbf{Video processing.}  In the domain of video and audio, multimodal fusion is also a growing trend. With the migration of image-text multimodal models to video-text and audio-text multimodal domains, a series of representative models have emerged. For example, the VideoCoCa model \cite{yan2022video} for the image-text domain. The CLIP model led to the development of the VideoCLIP model \cite{xu2021videoclip}. The advent of unified multimodal large models has also driven advancements in the field of video processing. Alibaba's mPLUG-2 \cite{xu2023mplug} has shown impressive performance in video-related tasks, e.g., video question answering and video captioning. Moreover, Google's MusiclM \cite{agostinelli2023musiclm} has gained recognition in the audio multimodal domain, as it can generate music based on text inputs. In addition, the video and audio domains involve a range of other multimodal tasks. Audio-visual speech recognition is the task of performing speech recognition on given videos and audio of individuals. Video sound source separation involves localizing and separating multiple sound sources in a given video and audio signal. Image generation from audio refers to generating images related to given sounds. Speech-conditioned face generation involves generating videos of a speaking person based on given speech utterances. There are some tasks like audio-driven 3D facial animation, which can generate 3D facial animations of a speaking person based on a given speech, and a 3D facial template \cite{richard2021meshtalk}.

\textbf{Smarter digital human.}  AIGC technologies \cite{wu2023ai} have played an important role in the development of digital humans, simplifying the process and enhancing development efficiency. Companies like Meta and NVIDIA have introduced products to assist users in creating 3D digital humans, with NVIDIA's Omniverse Avatar being an example. Users can create digital humans by uploading photos, videos, or audio, offering the advantages of efficiency and cost-effectiveness. Specifically, natural language generation technology impacts the quality of content in human-computer interactions, while computer vision technology affects the facial expressions and body movements of digital humans, such as lip synchronization \cite{prajwal2020lip}. The continuous advancement of AIGC technologies enables high-quality human-computer interactions. AIGC empowers AI-driven digital humans with intelligent development, providing recognition, perception, analysis, and decision-making capabilities during multimodal interactions.

\textbf{Practical guide for data.} Multimodal datasets play a crucial role in advancing research on vision and language tasks. These datasets combine different modalities, such as images, text, videos, and audio, providing rich and diverse sources of information for various applications. We categorize the multimodal datasets into different types and present a curated selection of representative datasets for each category, as shown in Table \ref{table3}. For future research, we can use these datasets to conduct experiments to test the model's effectiveness.

\begin{table}[h]
    \caption{The multimodal datasets}
    \begin{tabular}{|c|c|c|c|c|}
    \hline
    \textbf{Datasets}  & \textbf{Year}  & \textbf{Scale} & \textbf{Modalities} & \textbf{Paper}  \\
    \hline
    COCO  & 2014 & 567K & Image-Text &  \cite{lin2014microsoft} \\
    \hline
    Visual Genome  & 2017 & 5.4M & Image-Text &  \cite{krishna2017visual} \\
    \hline
     YouCook2 & 2018 & 2.2K & Video-Text  &   \cite{zhou2018towards} \\
    \hline
      WebVid2M & 2021 & 2.5M & Video-Text  &   \cite{bain2021frozen} \\
      \hline
      Common Voice & 2019 &  9.2K & Audio-Text  &   \cite{ardila2019common} \\
        \hline
      LibriSpeech & 2015 & 1K & Audio-Text  &   \cite{panayotov2015librispeech} \\
    \hline
      M5Product & 2021 & 6M & Image-Text-Video-Audio  & \cite{dong2021m5product} \\
    \hline
     MSR-VTT & 2016 & 10K & Image-Text-Video-Audio  & \cite{xu2016msr} \\
    \hline
    \end{tabular}
    \label{table3}
\end{table}

\section{Challenges} \label{sec:challenges}

To further improve the performance of multimodal applications, some fundamental issues still require more attention, including but not limited to:

\textbf{Modalities expansion.} The sensors and data sources are diverse, so they can acquire rich information in order to achieve more comprehensive and accurate analysis and recognition. For example, in the field of emotion computation, modality expansion involves using multiple modalities such as audio, facial expressions,  electrocardiography (ECG), and electroencephalography (EEG) to gain a more comprehensive understanding and recognition of people's emotional states \cite{katsigiannis2017dreamer}. The audio modality can capture changes in the speaker's tone and speech rate; the visual modality can analyze facial expressions and body language; and the ECG and EEG can provide physiological signals related to emotional changes. In addition, the field of medical imaging involves multiple modalities such as CT scans, MRIs, and PET. For example, CT scans can provide detailed information about tissue structure and lesions; MRI can observe the anatomical structures and functionality of tissues; and PET can be used to detect metabolism and the distribution of biomarkers. By combining different modalities of image data, doctors, and researchers can obtain more comprehensive and accurate medical information to support precise diagnosis and treatment decisions.

\textbf{Time-consuming problem.} For optimizing training architectures and improving training time, large models have a significant impact on AI systems. Firstly, due to the models' enormous scale, computations may need to be distributed across clusters. Secondly, multi-user and multi-task scenarios are common, requiring support for multi-tenancy. Moreover, high reliability is essential, demanding models to have dynamic fault tolerance capabilities. Multiple backbone models need to be combined. While multimodal LLMs have achieved tremendous success in various domains, their computational requirements pose significant challenges to model training. How can we accelerate model training \cite{zeng2023distributed}? We can dynamically allocate multiple models of different architectures to two high-speed interconnected data centers. During training and inference, pathways dynamically schedule models through gang scheduling, enabling capabilities such as shared computation, shared weights, and dynamic routing \cite{driess2023palm}.

\textbf{Lifelong/continual learning}. The current classic approach is to run an AI algorithm on a given dataset, build a model, and then apply this model to an actual task. This is called isolated learning and causes the shortcoming that the algorithm does not have memory capabilities. Therefore, the model or algorithm does not retain the learned knowledge and then continually apply it to future learning. For real applications but not an isolated task, multimodal large models require the ability of lifelong learning \cite{chen2018lifelong} or continual learning \cite{zenke2017continual}. We should build an LLM with continuous learning capabilities that can make a complex understanding of the world based on its own experience, thereby using more complex knowledge for autonomous and progressive training and improvement \cite{zenke2017continual}.

\textbf{Towards AGI.} On the path toward artificial general intelligence (AGI), we still face many opportunities and challenges. For example, the catastrophic forgetting problem \cite{chen2018lifelong} refers to the phenomenon where a neural network and its associated weights, originally trained for a language task, are repurposed for other tasks, resulting in the network forgetting its initial training objectives. In such cases, the large model may lose its original language capabilities, leading to a decline. For example, in language ability when shifting to robotic-based applications \cite{zeng2023large}. Recent research like BLIP-2, KOSMOS-1, BEiT-3, and PaLI \cite{chen2022pali} has highlighted two feasible approaches to address this issue: i) avoid catastrophic forgetting by using smaller networks and retraining from scratch with new data; ii) circumvent catastrophic forgetting by employing larger language networks as backbones. Note that there are still other challenges when pursuing AGI, including multimodal fusion, multimodal alignment, co-learning, and model-as-a-service (MaaS) \cite{gan2023model}.

\section{Conclusion}  \label{sec:conclusion}

The advancements in multimodal models have opened up new avenues for AI, which enables binary machines to understand and then process diverse data types. Multimodal models will lead to more comprehensive and intelligent systems in the near future. We have provided a comprehensive exploration of multimodal model development. We first introduced the multimodal concept and then sorted out the historical development of multimodal algorithms. After that, we discussed the efforts of major technology companies in developing multimodal products and offered insights into the technical aspects of multimodal models. We also presented a compilation of commonly used datasets that can provide valuable experimentation and evaluation resources. Finally, the challenges associated with the development of multimodal models were highlighted and discussed for further research. By addressing these aspects, this paper aims to provide a deeper understanding of multimodal models and their potential characters in various domains.

\section*{Acknowledgment}
This research was supported in part by the National Natural Science Foundation of China (Nos. 62002136 and 62272196), and the Young Scholar Program of Pazhou Lab (No. PZL2021KF0023). Dr. Wensheng Gan is the corresponding author of this paper.

\bibliographystyle{IEEEtran}
\bibliography{mmmodel.bib}

\end{document}